# HieroLM: Egyptian Hieroglyph Recovery with Next Word Prediction Language Model

**Xuheng Cai**
Department of Computer Science
Stanford University
xuheng@stanford.edu

**Erica Zhang**
Department of Management Science
and Engineering
Stanford University
yz4232@stanford.edu

| | |
|---|---|
| Hieroglyphs | 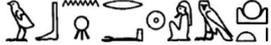 |
| Transliteration | wbn rꜥ m ꜣḫt |
| Transliteration (MdC) | wbn ra m Axt |
| English Translation | Re (the Sun God) rises in the horizon. |

Table 1: An example of transliteration and translation of a hieroglyphic sentence.

## Abstract

Egyptian hieroglyphs are found on numerous ancient Egyptian artifacts, but it is common that they are blurry or even missing due to erosion. Existing efforts to restore blurry hieroglyphs adopt computer vision techniques such as CNNs and model hieroglyph recovery as an image classification task, which suffers from two major limitations: (i) They cannot handle severely damaged or completely missing hieroglyphs. (ii) They make predictions based on a single hieroglyph without considering contextual and grammatical information. This paper proposes a novel approach to model hieroglyph recovery as a next word prediction task and use language models to address it. We compare the performance of different SOTA language models and choose LSTM as the architecture of our HieroLM due to the strong local affinity of semantics in Egyptian hieroglyph texts. Experiments show that HieroLM achieves over 44% accuracy and maintains notable performance on multi-shot predictions and scarce data, which makes it a pragmatic tool to assist scholars in inferring missing hieroglyphs. It can also complement CV-based models to significantly reduce perplexity in recognizing blurry hieroglyphs. Our code is available at https://github.com/Rick-Cai/HieroLM/.

## 1 Introduction

Egyptian hieroglyphs is the formal written language and an important medium for religious and funerary practices in Ancient Egypt. The process of decoding hieroglyphs involves first converting them into transliterations and then translating the transliterations into modern languages (Gardiner, 1927). Table 1 presents an illustration of this decoding process on a sample hieroglyphic sentence.

Due to natural erosion, it is common that the hieroglyphs on the surface of the unearthed artifacts are blurry or even missing. Efforts have been made to assist the process of recognizing blurry hieroglyphs with computer vision (CV) -based techniques (Barucci et al., 2021, 2022; Aneesh et al., 2024). Specifically, these works formulate hieroglyph recognition as an image classification task and use CV models such as convolutional neural networks (CNNs) to classify the blurry symbols. However, there are two major limitations in the CV-based approaches: **(i)** They cannot handle severely damaged or completely missing hieroglyphs because they rely on the visual characteristics of the signs. **(ii)** They make predictions based on a single hieroglyph, without considering the contextual and grammatical information contained in surrounding words that could help narrow down possibilities and significantly reduce perplexity.

As an example, the blurry hieroglyph A in the blue box in Figure 1 would confuse a CV model, because it could be either ⳁ (nḫb) or ⳁ (sw) based on its vague shape, but from the surrounding words we know that this sentence describes an offering by the king to the god Osiris, so it is likely that this blurry sign is ⳁ (sw), which means "the king". Moreover, for the red box in Figure 1, the signs are almost entirely missing, and the CV models will become useless, but from the words before it, we know that it should be a title of Osiris, which indicates that the missing word is probably 𓂦𓅓 (ḏdw), because ⌢𓂦𓅓 (nb ḏdw; "lord of Djedu") is a common title for Osiris in the offering formula.

In light of these limitations, we propose a novel approach where we model hieroglyph recovery as a next word prediction problem, which can be addressed effectively with language models. To select the best architecture for our task, we consider the following characteristics of Egyptian hieroglyphs (Allen, 2000): (i) It is a dead language whose corpora have ceased to grow, and thus the amount of data available for training is very limited. Hence, our model must be comfortable with small-scale training data. (ii) In Ancient Egypt, hieroglyphs are mostly used in limited scenarios including funerals, religious rituals, and monumental inscriptions. The restrictive formats on the hieroglyphic sentences leads to a better hope of accurate language modeling with simpler architectures. (iii) Due to its limited scope of usage, the hieroglyphic sentence structure has strong local affinity (e.g., a large portion of a sentence could be titles following names of gods or kings), suggesting that our model should have strong capability in capturing short-range dependencies. Based on these characteristics, we build our HieroLM with LSTM (Hochreiter and Schmidhuber, 1997). To validate our design choice, we compare the performance of HieroLM with popular architectures such as RNN (Medsker and Jain, 1999) and Transformers (Vaswani et al., 2017) in Section 4.3.

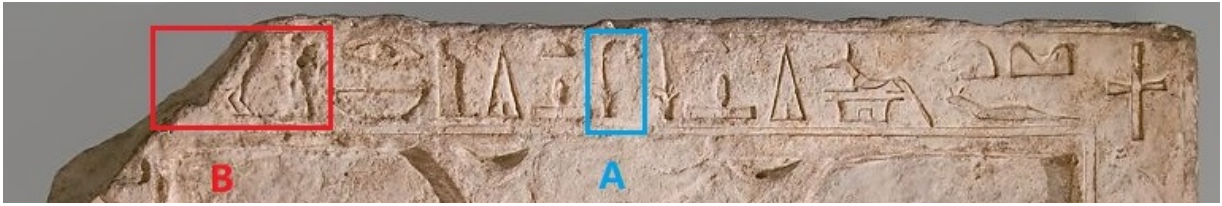

Figure 1: A Middle Kingdom tablet at The Metropolitan Museum of Art.[1] Hieroglyph A in the blue box is an example of blurry hieroglyphs. Hieroglyph B in the red box is an example of (nearly) missing hieroglyphs.

Our contributions are summarized as follows:

- To the best of our knowledge, this is the first paper to model hieroglyph recovery as a next word prediction task addressed with language models.
- We propose HieroLM based on LSTM, which achieves over 44% accuracy (i.e., it infers missing words correctly almost half of the time).
- Experiments show that HieroLM is robust enough to maintain notable performance on both multi-shot prediction and scarce context.

## 2 Related Work

### 2.1 Hieroglyph Recognition with CV

Modeling hieroglyph recognition as an image classification task is well-explored. Franken et al. (Franken and van Gemert, 2013) proposed to use the Histogram of Oriented Gradients (HOG) and the Shape-Context (SC) descriptors to extract and compare hieroglyphs. The HOG method was later enhanced with Region of Interest (ROI) extraction (Elnabawy et al., 2021). Moustafa et al. (Moustafa et al., 2022) and Aneesh et al. (Aneesh et al., 2024) explored the performance of ShuffleNet, MobileNet, ResNet, VGG, DenseNet, and Inception v3 on hieroglyph recognition, while Glyphnet (Barucci et al., 2021) achieves the state-of-the-art performance. However, these CV models rely heavily on the visual quality of the signs and fail to incorporate contextual information.

### 2.2 Next Word Prediction with LMs

Next-word prediction involves predicting the subsequent word in a sequence given the preceding context. Early approaches use n-gram models which suffer from data sparsity and limited context understanding. NPLM (Bengio et al., 2000) addresses the limitations of n-gram models with neural networks. CSLM (Schwenk, 2007) projects words to a continuous space to handle variable-length contexts. Recurrent neural networks (RNNs) and long short-term memory (LSTM) (Hochreiter and Schmidhuber, 1997) greatly improve the prediction accuracy with recurrent model architectures to maintain memory and capture local dependencies. Recently, Transformers (Vaswani et al., 2017) revolutionizes language modeling by employing self-attention to consider the entire input context, but it is less-suited for our task due to the limited data availability.

## 3 Methodology

In this section, we describe in detail our HieroLM model, which adopts the LSTM architecture as illustrated in Figure 2.

Assume that the input sentence has $T$ words. Let $x^{(t)} \in \{0,1\}^{|V|}$ be the one-hot encoding of the $t$-th word ($1 \leq t \leq T$) in the sentence. Then, its embedding $e^{(t)} \in \mathbb{R}^s$, where $s$ is the embedding size, is computed as $e^{(t)} = Ex^{(t)}$, where $E$ is an embedding layer. The hidden state $h^{(t)} \in \mathbb{R}^d$, where $d$ is the hidden dimension size, at step $t$ is computed as:

$$h^{(t)} = \mathcal{F}_\theta(h^{(t-1)}, e^{(t)})$$

where $\mathcal{F}_\theta$ incorporates long short-term memory (Hochreiter and Schmidhuber, 1997). Specifically, given $h^{(t-1)}$ and $e^{(t)}$, we compute the following states with single layer neural networks:

$$f^{(t)} = NN_{forget}(h^{(t-1)}, e^{(t)})$$
$$i^{(t)} = NN_{in}(h^{(t-1)}, e^{(t)})$$
$$g^{(t)} = NN_{gate}(h^{(t-1)}, e^{(t)})$$
$$o^{(t)} = NN_{out}(h^{(t-1)}, e^{(t)})$$

---

[1] Source: https://www.metmuseum.org/art/collection/search/545055.

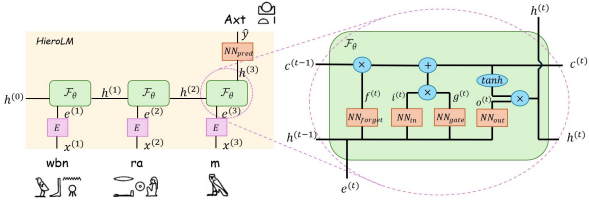

Figure 2: Model structure of HieroLM.

The cell state $c^{(t)} \in \mathbb{R}^d$ at step $t$ is computed as:

$$c^{(t)} = f^{(t)} \odot c^{(t-1)} + i^{(t)} \odot g^{(t)}$$

where $c^{(0)}$ is the initial cell state. Finally, the hidden state $h^{(t)}$ is calculated as:

$$h^{(t)} = o^{(t)} \odot tanh(c^{(t)})$$

The predicted output is calculated by:

$$\hat{y} = NN_{pred}(h^{(T)})$$

where $NN_{pred}$ is a single neural layer plus a softmax layer, which projects the final hidden state from $d$ to the size of the vocabulary $|V|$.

## 4 Experiments

### 4.1 Datasets

We evaluate our model and the baselines on three real-world datasets with hieroglyphic sentences from unearthed Egyptian artifacts. The dataset statistics are summarized in Table 2.

- *Ancient Egyptian Sentences (AES)* (Jauhiainen and Jauhiainen, 2023): It is a collection of over 100,000 ancient Egyptian sentences across multiple dynasties.

- *The Ramses Transliteration Corpus* (Rosmorduc, 2020): It contains transliterations converted from a large corpus of Late Egyptian sentences.

- *Mixed*: Since AES contains sentences from different eras while texts in Ramses come from Late Egypt, they have different distributions due to language evolution. To evaluate the models' cross-distribution modeling ability, we synthesize AES and Ramses into a mixed dataset.

We use the MdC transliterations of the hieroglyphic sentences throughout our experiments because it replaces irregular letters (e.g., ʿ and ꜣ) in the common transliteration with English letters (e.g., "a" and "A") for convenient processing. The sentences are split into training, validation, and test sets by an 8:1:1 ratio.

Table 2: Dataset statistics.

| Dataset | Sentence # | Vocab # | Training # | Validation # | Test # |
| --- | --- | --- | --- | --- | --- |
| AES | 98,375 | 7,058 | 78,801 | 9,800 | 9,774 |
| Ramses | 61,069 | 3,499 | 48,848 | 6,116 | 6,105 |
| Mixed | 159,444 | 8,436 | 127,649 | 15,916 | 15,879 |

### 4.2 Baselines

We compare our LSTM-based HieroLM model with the following widely-adopted baselines:

- *Neural Probabilistic Language Model (NPLM)* (Bengio et al., 2000). We use a trigram NPLM as the naivest baseline.

- *Recurrent Neural Network (RNN)* (Medsker and Jain, 1999). We adopt a unidirectional, single-layer RNN. This also serves as an ablated version of HieroLM where the long short-term memory is removed.

- *Transformer* (Vaswani et al., 2017). We employ a single-layer encoder with `nheads=16` and `dropout = 0` due to limited data.

### 4.3 Performance Validation

We summarize the main results in Table 3, with the following observations:

- **Hieroglyphic vocabulary is restrictive.** Next word prediction is intrinsically hard due to the high degree of freedom of modern languages. There are often multiple legitimate next words that make perfect grammatical and semantic senses for an input context. The SOTA LSTM-based language model for English trained on billion-scale datasets by Google only achieves a perplexity of 30 (Jozefowicz et al., 2016). However, HieroLM achieves a perplexity of ~26 with less than a million words, indicating that the hieroglyphic vocabulary is highly constrained.

| Dataset | Metric | NPLM | Transformer | RNN | **HieroLM** |
|---|---|---|---|---|---|
| AES | Perplexity | 41.57 | 52.21 | 42.25 | **26.50** |
| | Accuracy | 0.3075 | 0.3143 | 0.3828 | **0.4525** |
| | F1 Score | 0.0485 | 0.0488 | 0.1201 | **0.1420** |
| Ramses | Perplexity | 28.75 | 38.59 | 31.89 | **21.59** |
| | Accuracy | 0.3553 | 0.3727 | 0.4387 | **0.4895** |
| | F1 Score | 0.0775 | 0.0905 | 0.1933 | **0.2074** |
| Mixed | Perplexity | 42.14 | 53.78 | 43.34 | **26.48** |
| | Accuracy | 0.3022 | 0.3151 | 0.3801 | **0.4450** |
| | F1 Score | 0.0481 | 0.0466 | 0.1377 | **0.1421** |

Table 3: Main performance results.

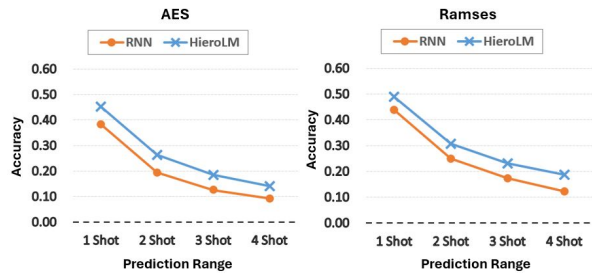

Figure 3: Multi-shot prediction accuracy.

- **Recurrent architecture dominates.** As the table shows, in face of small datasets, models with recurrent architecture (HieroLM and RNN) exhibit consistent superiority. This proves the recurrent models' ability to capture local semantic affinity with limited data.

- **LSTM enhances performance.** The comparison between HieroLM and RNN is a natural ablation study. The outperformance of HieroLM w.r.t. RNN proves that LSTM can enhance the model by long-range perception.

- **Transformer is less-suited for this task.** We can see that Transformer underperforms HieroLM, which demonstrates that its architecture is less suitable for this task due to limited data.

### 4.4 Multi-shot Prediction Performance

In reality, it is common for a number of contiguous hieroglyphic words to be missing together, which makes it important to evaluate the model's ability to predict a series of words accurately without teacher forcing. Figure 3 presents the accuracy of HieroLM for multiple following words. We can observe a favorable diminishing decrease in accuracy with the increase of prediction range. It is also worth noting that the model maintains an accuracy of over 14% on predicting 4 words in a row.

### 4.5 Resistance against Data Scarcity

A big obstacle in leveraging ML for hieroglyph recovery is data scarcity, which manifests on two levels: On the corpus level, the total number of hieroglyphic sentences from ancient artifacts are limited. On the sentence level, many hieroglyphic sentences are incomplete due to erosion, with only few identifiable symbols. The short context increases difficulty in inferring missing words. To evaluate HieroLM's robustness in resisting the sentence-level data scarcity, we group test sentences by their length and calculate accuracy of HieroLM and RNN on each group. Figure 4 shows that the models generally perform worse with shorter context (except group [1,5) on AES, as AES contains many short but formulaic phrases), but HieroLM consistently outperforms RNN on all context lengths, demonstrating its robustness under scarce input.

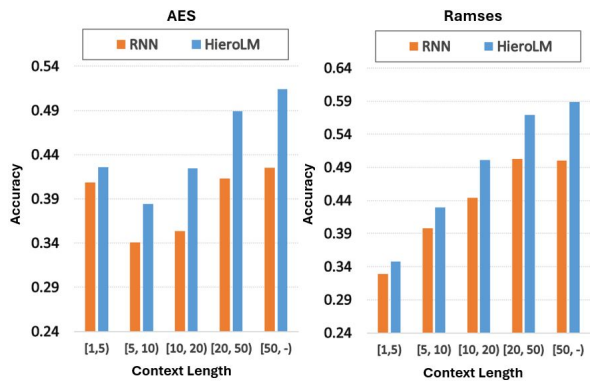

Figure 4: Accuracy with different context lengths.

### 4.6 Word Embedding Quality

In this section, we illustrate the effectiveness of HieroLM by inspecting the word embeddings it learns on the Mixed dataset. Specifically, we map the embeddings of all words to the 2-D space with PCA and visualize some common words that frequently appear on Egyptian artifacts in Figure 5, which shows a distribution of word embeddings that reflects the semantic of *offering* from the subjects (the mortals) to the targets (the gods).

### 4.7 Hyperparameter Analysis

We explore the sensitivity of HieroLM with respect to key hyperparameters including embedding size, hidden dimension size, and dropout rate. The results also provide ground for our choice of hyperparameters. Due to space limit, we present the results in Appendix B.

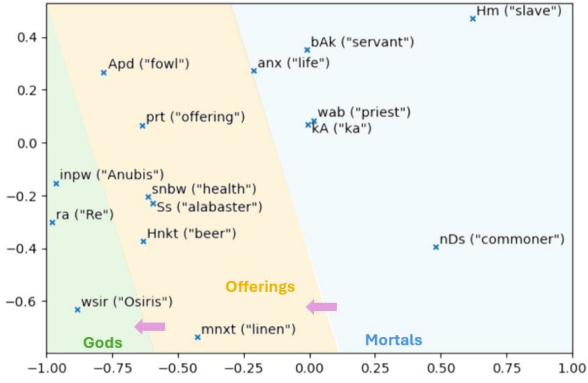

Figure 5: Embeddings of common hieroglyphic words.

### 4.8 Case Study

We demonstrate HieroLM's ability to learn semantic patterns by two concrete cases corresponding to two common patterns in Egyptian hieroglyphs.

*Case 1: Offering formula.* Below is the #1563 sentence in the test set of the Mixed dataset.

Processed MdC:

n kA n | wr swn w | pnTw | mAa xrw

Transliteration:

n k3 n | wr-swn.w | pntw | m3ˁ ḥrw

English Translation:

For the *ka* of | the great physician | Pentu , | the true of voice. [2]

This sentence is a common conclusion of the offering formula. It has a fixed format: [n k3 n] + [*Title and name of the deceased*] + [m3ˁ ḥrw], where m3ˁ ḥrw ("the true of voice") is a universal title for the deceased. Upon seeing n k3 n and the title and name of the deceased, HieroLM is capable of predicting that the following words are m3ˁ ḥrw. Note that this is an example of successful 2-shot prediction.

*Case 2: Titles of kings.* Below are the first few words of #8779 sentence in test set of the Mixed dataset.

Processed MdC:

nswt bj tj | nb tA du | wsr mAa t raw stp n | jmn | zA ra ...

Transliteration:

nswt-bity | nb t3.du | wsr-m3ˁt-rˁ stp.n-imn | s3 rˁ ...

English Translation:

King of Upper and Lower Egypt, | Lord of the Two Lands, | Ramesses IV, | Son of Re ...

---

[2] In ancient Egypt, *ka* refers to a part of human soul that leaves the body upon death.

This part of the sentence contains the name and titles of the king Ramesses IV. Titles of kings in ancient Egypt have rigorous formats. nswt-bity ("King of Upper and Lower Egypt") is the title preceding the coronation name of the king, and s3 rˁ ("Son of Re") is a title commonly following the king's name. After seeing nswt-bity and the name of the king, HieroLM can infer that the following words are likely to be s3 rˁ. When we feed in the sequence "nswt bj tj nb tA du wsr mAa t raw stp n jmn", the model responds with "zA", and when appending "zA" to the input, it outputs "ra", which is also a 2-shot prediction example.

## 5 Conclusion

In this paper, we exclusively propose to model Egyptian hieroglyph recovery as a next word prediction task addressed by language models. Considering the data scale and the local semantic affinity, we propose HieroLM with LSTM architecture, which achieves significant accuracy in experiments. Its notable performance on multi-shot predictions and short input contexts makes it practical in archaeological research to infer missing hieroglyphs and complement CV models. In the future, we plan to explore potential ways of integrating computer vision models and language models into a unified and effective hieroglyph recovery system.

## 6 Limitations

In this work, due to limited data availability, we had little success in leveraging the power of the state-of-the-art Transformer models. While it is not impossible to tailor Transformer to smaller-scale data, it requires sophisticated training techniques (Popel and Bojar, 2018) and is known to be less robust in some cases (Liu et al., 2022). In the future, we aim to explore how self-attention-based architectures can be adapted to Egyptian hieroglyphic texts.

## 7 Acknowledgement

We sincerely appreciate the guidance and help from Prof. Christopher Manning and Ms. Anna Goldie at Stanford NLP Group.

We are also grateful for the support from Prof. Jose Blanchet at Stanford Management Science and Engineering.

Special thanks to Mr. Amjad Refai at University of Hong Kong for Egyptology knowledge.

## A  More Details on Experimental Settings

### A.1  Evaluation Metrics

We evaluate the models on 3 metrics:

- *Perplexity.* It measures the model's probability of predicting the correct word. A lower perplexity score indicates better predictive performance and a higher confidence for the prediction.

- *Accuracy.* It is the ratio between the number of correct predictions and the total predictions. It reflects the practical efficacy of our models in real-world application.

- *F1 Score.* This metric harmonizes precision and recall, providing a balanced view of performance across all classes. We use the macro averaging method in F1 calculation.

### A.2  Hyperparameters and Training Configurations

For fair comparison, we adopt an embedding size of 1024 and a hidden dimension size of 1024 for HieroLM and all the baselines, based on the hyperparameter analysis in Section 4.7. The dropout rate is searched individually for each dataset. We employ a learning rate decay and early stopping strategy, such that when the validation perplexity stops decreasing for 5 epochs, the learning rate decays by half, and the training will be stopped after five decays.

## B Hyperparameter Analysis

In this section, we investigate the sensitivity of HieroLM with respect to key hyperparameters including embedding size, hidden dimension size, and dropout rate. The results, as summarized in Figure 6, also provide basis for our choice of hyperparameters.

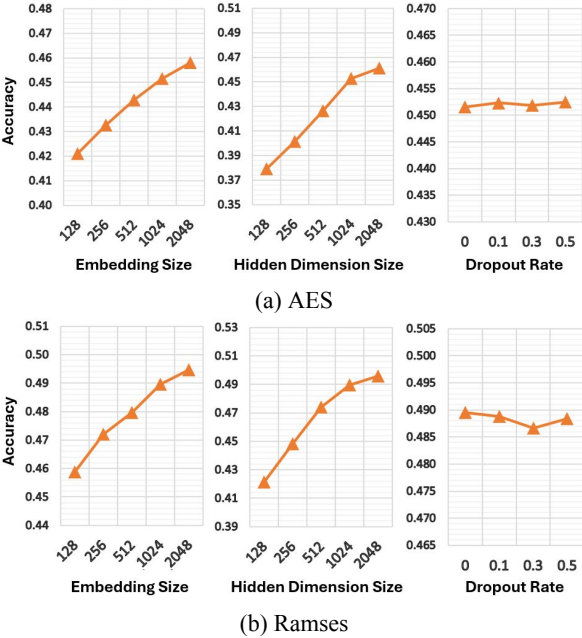

(a) AES

(b) Ramses

Figure 6: Test accuracy w.r.t. embedding size, hidden dim size, and dropout rate.